\documentclass[runningheads]{llncs}
\usepackage[T1]{fontenc}
\usepackage[utf8]{inputenc}

\usepackage{amsmath}

\usepackage{graphicx}
\usepackage{listings}
\usepackage{xcolor}

\usepackage[hidelinks]{hyperref}
\usepackage{url}

\usepackage{floatrow}
\newfloatcommand{capbtabbox}{table}[][\FBwidth]

\newcommand{\ws}[1]{\lstinline|#1|} 
\newcommand{\up}[1]{\lstinline|#1|}

\newcommand{\prprgraph}[0]{$\textit{Pr}^2\textit{Graph}$}

\definecolor{codegreen}{rgb}{0,0.6,0}
\definecolor{codegray}{rgb}{0.5,0.5,0.5}
\definecolor{codepurple}{rgb}{0.58,0,0.82}
\definecolor{backcolour}{rgb}{0.95,0.95,0.92}

\lstdefinestyle{mystyle}{
    backgroundcolor=\color{backcolour},   
    commentstyle=\color{codegreen},
    keywordstyle=\color{magenta},
    numberstyle=\tiny\color{codegray},
    stringstyle=\color{codepurple},
    basicstyle=\ttfamily\footnotesize,
    breakatwhitespace=false,         
    breaklines=true,                 
    captionpos=b,                    
    keepspaces=true,                 
    numbers=left,                    
    numbersep=5pt,                  
    showspaces=false,                
    showstringspaces=false,
    showtabs=false,                  
    tabsize=2
}

\begin{document}

\title{An LLM-enabled semantic-centric framework to consume privacy policies}
%
%
\author{Rui Zhao\inst{1}\orcidID{0000-0003-2993-2023} \and
Vladyslav Melnychuk\inst{1} \and
Jun Zhao\inst{1}\orcidID{0000-0001-6935-9028} \and Jesse Wright\inst{1}\orcidID{0000-0002-5771-988X} \and Nigel Shadbolt\inst{1}\orcidID{0000-0002-5085-9724}}
\authorrunning{R. Zhao et al.}
%
\institute{University of Oxford, Oxford, UK\\
\email{\{rui.zhao,jun.zhao\}@cs.ox.ac.uk}\\\email{\{jesse.wright,nigel.shadbolt\}@jesus.ox.ac.uk}\\
\email{vladyslav.melnychuk18@gmail.com}}
\maketitle              
\begin{abstract}
In modern times, people have numerous online accounts, but they rarely read the Terms of Service or Privacy Policy of those sites, despite claiming otherwise, due to the practical difficulty in comprehending them. The mist of data privacy practices forms a major barrier for user-centred Web approaches such as Solid, and for data sharing and reusing in an agentic world.
Existing research proposed methods for using formal languages and reasoning for verifying the compliance of a specified policy, as a potential cure for ignoring privacy policies. However, a critical gap remains in the creation or acquisition of such formal policies at scale.
We present a semantic-centric approach for using state-of-the-art Natural Language Processing (NLP) tools, namely large language models (LLM), to automatically identify key information about privacy practices from privacy policies, and construct \prprgraph{}, knowledge graph with grounding from Data Privacy Vocabulary (DPV) for privacy practices, to support downstream tasks.
Along with the pipeline, the \prprgraph{} for the top-100 popular websites is also released as a public resource, by using the pipeline for analysis. We also demonstrate how the \prprgraph{} can be used to support downstream tasks by constructing formal policy representations such as Open Digital Right Language (ODRL) or perennial semantic Data Terms of Use (psDToU).
To evaluate the technology capability, we enriched the Policy-IE dataset by employing legal experts to create custom annotations. We benchmarked the performance of different large language models for our pipeline and verified their capabilities.
Overall, the pipeline and relevant resources shed light on the possibility of large-scale analysis of online services' privacy practices, as a promising direction to audit the Web and the Internet.
We release all datasets and source code as public resources to facilitate reuse and improvement.

\end{abstract}

\section{Introduction}

Collectively, we are subscribing to an ever-increasing number of online services - each of which has us sign custom ``Terms of Service'' or ``Privacy Policies'' to enable the collection and use of our data. 
Despite the privacy and legal implications, less than 7\% of users read these agreements, making them ``the biggest lie on the Internet''~\cite{obar_biggest_2020}.
A major cause for that is information overload, as the average privacy policy requires a 29-minute read time. 
Current regulations, including the General Data Protection Regulation (GDPR)~\cite{councilofeuropeanunion_regulation_2016} and the Digital Service Act (DSA)~\cite{councilofeuropeanunion_regulation_2022}, exacerbate the issue, requiring companies to collect more permissions from users without providing technical or legal standards for facilitating users, thus more burdens for the users.


As a reaction to the online privacy challenges, decentralized personal data stores \cite{fallatah_personal_2023} have emerged as an alternative pattern: users hold data in their own storage and manage data access, while applications/services make requests to use the data. In particular, Solid (Social Linked Data) \cite{sambra_solid_2016} pioneers using semantic technologies for data representation, for maximizing data interoperability and data reusing.

However, less discussion has been focused on \emph{how to support users to make decisions} regarding the data access and usage requests from the decentralized applications, thus still relying on privacy policies. This poses challenges for Solid towards an ethical Web ecosystem, where user-centric data governance, cross-boundary data sharing and intelligent autonomous agents all require well-understood data usage repercussions during (automated or manual) decision making.

Research on usage control \cite{breaux_eddy_2014,sandhu_usage_2003}, especially ODRL \cite{_odrl_2018} and psDToU \cite{zhao_perennial_2024} building on semantic technologies, provides promising solutions to this issue, where formal policies and automated reasoning are employed to check the compliance of data usage, thus further facilitating decision-making. Work \cite{zhao_perennial_2024} has also demonstrated the possibility of integrating a policy engine into Solid protocol, as well as supporting continuous (``perennial'') policy checking for downstream derived data. In addition, research such as DPV \cite{j.pandit_data_2025} also created standard interoperable vocabularies for expressing concepts in formal languages.

But an important question is not answered by such research: \textit{where do the formal policies come from}? Indeed, it would not be difficult to convert between different formal representations if a mapping is provided. Yet the problem lies in the source of information: at best, services only provide natural-language policies, rather than formal policies.



In this paper, we pioneer a software pipeline for automatically converting natural-language privacy policies into formal knowledge of the privacy practices in them, as a knowledge graph called \prprgraph{}, for the sake of interoperability and extensibility provided by RDF \cite{w3c_rdf_2014}. 
The pipeline takes advantage of advancements of large-language models (LLMs) \cite{wang_superglue_2019,hendrycks_measuring_2020,srivastava_imitation_2023}, for both their remarkable natural-language-understanding performance, and their wide availability -- off-the-shelf tools allow usage by people with modest technical abilities.

We argue the necessity of a KG-centric view, instead of the general view on complementing LLM/NLP with KG \cite{pan_unifying_2024}, in this context: unambiguous and precise meanings of policy terms are a necessity, while LLMs can only provide a probabilistic interpretation; issues such as hallucination \cite{huang_survey_2025} also reduce the appropriateness of LLMs; appropriate prompting engineering \cite{liu_pretrain_2023,sahoo_systematic_2025} is still necessary in the forseeable future when new queries are needed. On the other hand, KGs are extensible by design; through technologies like OWL \cite{w3cowlworkinggroup_owl_2012}, KGs provide the needed formality and interoperability. Their white-box nature also supports auditing of information, before downstream tasks consume them, or to inform changes.

Along with the pipeline, we also provide the \prprgraph{} for the top-100 most-visited websites, as an open and accessible information source, benefiting users with less technical or resource advantages; we also showcase a valuable downstream task for converting \prprgraph{} into formal policies of psDToU and ODRL powered by semantic technologies.
In addition, we constructed a dataset by using human annotators, extending the Policy-IE dataset \cite{ahmad_intent_2021}, to form a high-quality basis for evaluating NLP pipeline's accuracy -- despite being phenomenal, LLMs' capability in understanding privacy policies is not yet well-understood (Sec \ref{sec:background:pp-analysis}). We used this dataset to benchmark the performance of several state-of-the-art LLMs in different settings, verifying the capability of the models and thus the pipeline. 

Together, they form a complete story in providing, evaluating and demonstrating the utility of a semantic-centric NLP pipeline for privacy policy analysis and usage control, providing a valuable resource for filling in both practical and theoretical gaps of existing practices.

\subsection*{Resource Location}

Due to the nature of these resources, they are located separately for audiences with different targets:

\begin{enumerate}
    \item NLP pipeline and evaluation code: \url{https://github.com/renyuneyun/pp-analyzer}
    \item Privacy policy annotations: \url{https://doi.org/10.5281/zenodo.15392162}
    \item \prprgraph{} for top-100 most-visited websites\footnote{It also contains the formal policy in psDToU and ODRL as a demonstration of downstream tasks}: \url{https://doi.org/10.5281/zenodo.15408913}
\end{enumerate}



\section{Background and Related Work}
\label{sec:background}

\subsection{Privacy Policy Processing}
\label{sec:background:pp-analysis}

Privacy policies have been a research topic across several themes, with diverse methods and goals. Here, we present a review of the research on automatically identifying and re-presenting its main information, and those supporting users' decision-making, due to their close link to this work.

Different initiatives attempt to use summaritive and user-friendly labels, icons or badges to highlight the key privacy practices contained in the privacy policies or otherwise found in the software/service, such as Tos;DR (Terms of Service; Didn't Read)\footnote{https://tosdr.org/} and privacy (nutrition) labels or icons \cite{kelley_nutrition_2009,emami-naeini_ask_2020,efroni_privacy_2019}. However, two barriers exist for these approaches: 1) the labels need to be manually created; 2) only very limited information is contained in the labels.

Some work proposed self-trained NLP models to identify certain types of information from privacy policies, such as PolicyLint \cite{andow_policylint_2019}, Polisis \cite{harkous_polisis_2018} and PoliCheck \cite{andow_actions_2020}. They have different downstream tasks, such as creating privacy icons or detecting contradictions within the policies, thus being more generic than the previous category. However, they normally perform analysis on the lexical level with pre-defined patterns, without a clear way for others to reuse or customize. In contrast, PoliGraph \cite{cui_poligraph_2023} extracts information into a knowledge graph to facilitate further analysis. It supports richer semantics through the graph, but uses a task-specific graph structure, while not using standard or open vocabularies for the terms, thus limiting its generality to be extended and used by other people. In addition, all work in this category relies on custom self-hosted NLP models which can be a major burden to people without appropriate skills or resources.


With the development of large language models (LLMs) and public services providing easy access to them, some work attempted to evaluate their utility in analyzing privacy policies. In particular, \cite{savelka_unreasonable_2023}, LegalBench \cite{guha_legalbench_2023}, PolicyGPT \cite{tang_policygpt_2023} and \cite{rodriguez_large_2024} evaluated different LLMs' performances against certain types of queries using existing annotation datasets. They showed evidence of the advantages of using off-the-shelf LLMs for the privacy policy annotation tasks, thus incentivizing our design choice. However, they face the challenge of query tasks being simple, such as simply asking for boolean answers to the existence of different types of data practices, which both limits the types of supported queries, and incurs higher costs due to repeated queries.



Compared with existing work, this paper has two distinct features: 1. it is grounded in semantic technology and open standards, for better interoperability and easier auditing; 2. the query method is more scalable compared to existing LLM research with PPs, while also more accessible compared to custom NLP models.


\subsection{Privacy Policy Corpus}
\label{sec:background:pp-corpus}

Previous work has created several datasets for privacy policies, with different focuses.
Notably, the Usage Privacy Policy project \cite{sadeh_usable_2013} released multiple datasets, especially OPP-115, APP-350 and Privacy QA, as described below.

OPP-115 \cite{wilson_creation_2016} is a well-known early annotation dataset for 115 website privacy policies, and is widely used by later research. It contains annotations of nine types of data practices, each with further detailed questions, and the span of texts. Each annotation is performed on a given ``paragraph-length'' policy segment.
Likewise, APP-350 \cite{zimmeck_maps_2019} is a dataset of 350 mobile app privacy policies, focusing on the data type, the party, and modality.

PI-Extract \cite{bui_automated_2021} presented a 30-document dataset containing data types and practices (collecting or not, sharing or not). PrivacyQA \cite{ravichander_question_2019} (35 documents) and PolicyQA \cite{ahmad_policyqa_2020} (built on OPP-115) focused more on natural-language question-answering, where they collected questions related to privacy or policy concerns.

Policy-IE \cite{ahmad_intent_2021} is a distinct dataset that contains more detailed information about data practices, of 31 documents.
In addition to the data practices, it also contains granular information for them, especially party, action, (data and purpose) entity, and the \textit{relation} between them and the practices. It provides a good basis for advanced analysis due to its richer annotation.

However, none of the datasets provides or utilizes formal grounding of the concepts, thus suffering from issues of being unable to identify the same type of information, if their text uses different wordings. Therefore, based on Policy-IE, we create an enriched annotation dataset with such groundings using a standard vocabulary, DPV  \cite{j.pandit_data_2025} (Section \ref{sec:policy-ie-enrichment}).


\section{NLP Pipeline and \prprgraph{}}

The first resource we present is the software system, mainly the NLP pipeline, to convert natural-language privacy policies to a knowledge graph, \prprgraph{}. We also explain the structure of the \prprgraph{} in this section.

\subsection{NLP pipeline}

\begin{figure}
   \centering
   \includegraphics[width=.9\linewidth]{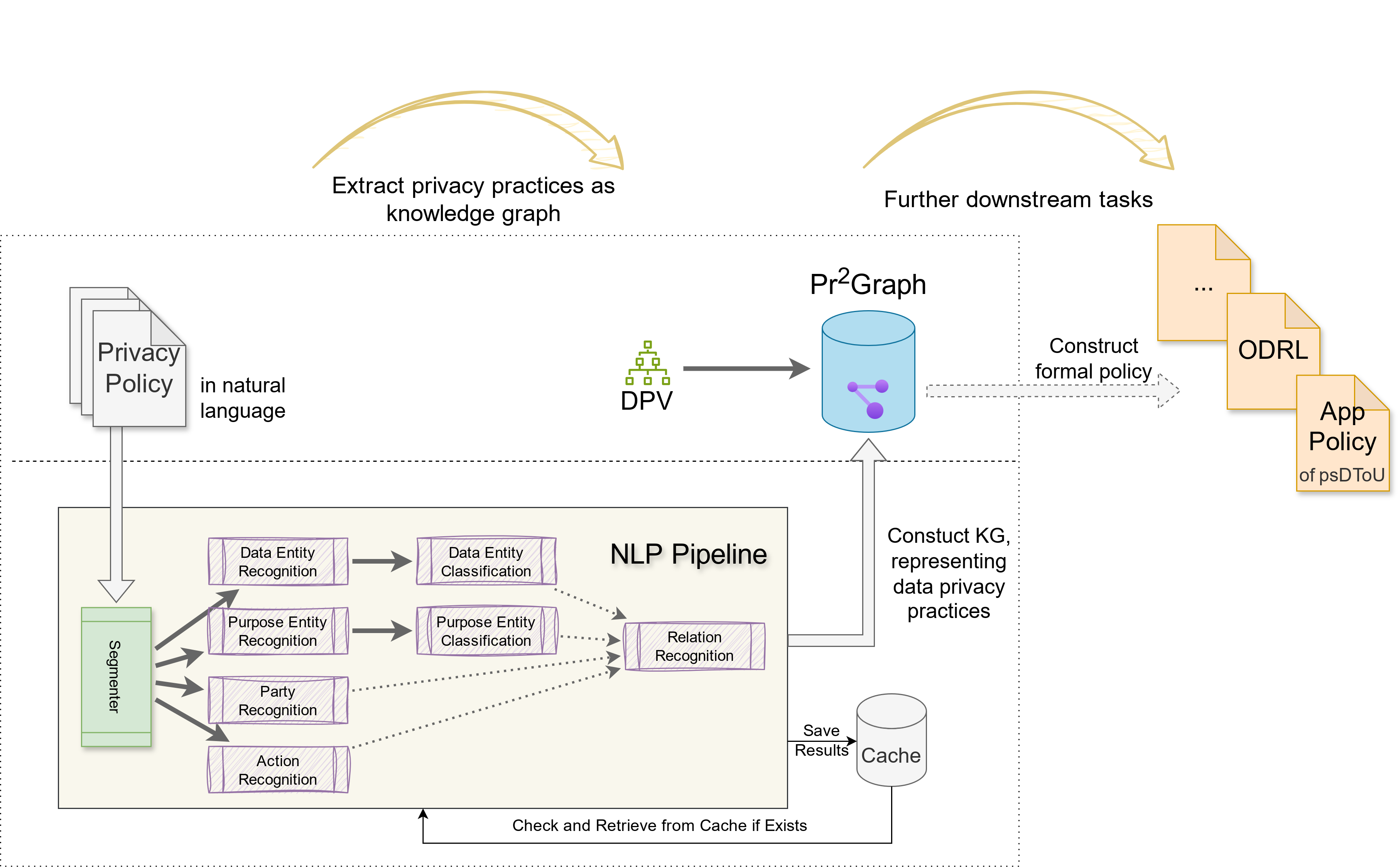}
   \caption{NLP pipeline, and downstream task}
   \label{fig:arch-pipeline}
\end{figure}

The \textit{NLP pipeline} is responsible for identifying relevant information from the privacy policy (PP), where the types of information are based on Policy-IE with additions (Sec \ref{sec:policy-ie-enrichment}). It is composed of several steps of different roles, as shown in Figure \ref{fig:arch-pipeline}. Their main tasks are:

\begin{itemize}
    \item Data entity recognition: identify the text spans from the PP that resemble data entities (e.g.~medical data);
    \item Purpose entity recognition: identify the text spans from the PP that resemble purpose entities (e.g.~advertizement purpose);
    \item Party recognition: identify the text spans from the PP that describes a party, and classify it as party type (e.g.~first-party and third-party);
    \item Action recognition: identify the text spans from the PP that describe the practice of data (e.g.~data collection, third-party data sharing, data storage retention, and data security protection);
    \item Data entity classification: classify the recognized data entity text to a canonical data entity in DPV;
    \item Purpose entity classification: classify the purpose entity text to a canonical purpose entity in DPV;
    \item Relation recognition: identify the relations between the previously recognized entities -- how the previously identified types of information relate to each other, especially what entities belong to what data practice, and serve what role (e.g.~the \textit{party} X is a \textit{data provider} for an \textit{action} Y).
\end{itemize}


Each step is performed as one query of the LLM\footnote{For our experiments, we used models from the gpt-4o family, but the system design is generic to all models.}, to a segment of the privacy policy, to identify the relevant information. In our current implementation, the segmentation is by line, which is empirically chosen to form a balance between cost-efficiency and accuracy.

Essentially, the tasks for Steps 1-6 are generally known as named entity recognition \cite{li_survey_2022}. Finally, all entities and data practices are grouped together by segment, and each segment is sent to perform relation extraction \cite{zhao_comprehensive_2024,wadhwa_revisiting_2023}. The results from these steps are then used to construct the data practices, and thus \prprgraph{}.

For data and purpose classification, the model will output the grounded terms in DPV, for better interoperability in further usage. Specifically, because purposes have a hierarchy, the model should predict the most accurate leaf node (subclass) in the hierarchy.

\subsubsection{Prompting strategy}

When invoking LLM predictions, in general, we use the system message to instruct the model to perform a specific task with a task description, and a description of the output (JSON) schema. We then provide the segment to analyze from the user prompt. Specifically, for the relation recognition task, we give each entity and practice (identified from previous steps) a unique ID from our code, and send them together in a user prompt; the task will ask the model to return the mapping as (a series of) tuples of (ID1, ID2, EventType).
Appendix \ref{appendix:llm-prompting-strategy} provides further details on the prompt design, including detailed prompt structure; exact prompt templates can be found in the source code.

\subsubsection{Output parsing}

Because LLMs do not always follow the instructions, the output can be different from the expected form. Examples of ill-formed responses include additional explanation, not giving JSON but a textual response, ill-formed JSON, and not using the said schema or keys. We perform appropriate post-processing before parsing the results, following heuristics discovered from response data, and existing helper libraries, especially \texttt{json-repair} \cite{jong_josdejong_2024}.

\subsection{\prprgraph{}: Privacy Practice Graph}

The parsed outputs are further composed and constructed as a knowledge graph, the \prprgraph{}, resembling the privacy practices described in PPs.

\begin{figure}
   \centering
   \includegraphics[width=.8\linewidth]{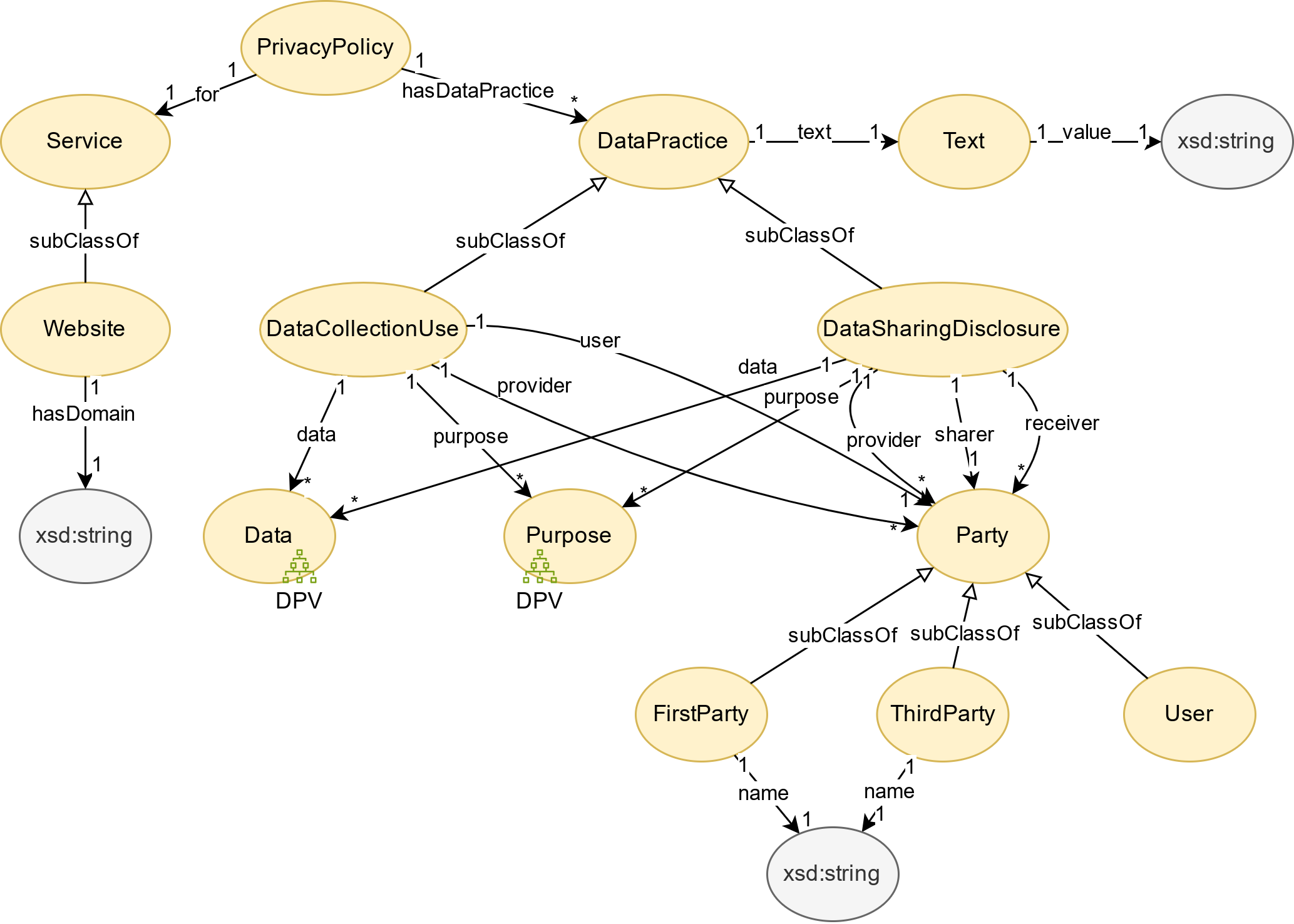}
   \caption{Overall classes and relations in \prprgraph{}.}
   \label{fig:pr2graph}
\end{figure}

Figure \ref{fig:pr2graph} shows the overall structure of the \prprgraph{}. The namespace for our introduced concepts is \lstinline|urn:pp-analyze:core#|. The graph is centred around data privacy practices \lstinline|DataPractice|, as it is the entity that links together data, purpose, parties, etc. It represents all information identified from the NLP pipeline:

\begin{itemize}
    \item Data usage practice
    \item Data type (in DPV) being used in a practice
    \item Purpose (in DPV) of a practice
    \item Party performing the practice
    \item Party providing the data
    \item Party receiving the shared data, if a data-sharing practice
    \item Original segment for each of the practices
\end{itemize}

Essentially, the collection of \lstinline|DataPractice| describes the flow of data: what data are involved, why are they used, who used them, and with whom they are shared. This covers practices like first-party data collection, third-party data collection (unified as \lstinline|DataCollectionUse|), and third-party data sharing, as commonly found in privacy policy annotations. But because terms are grounded with formal semantics, this representation possesses better interoperability, more suitable for automated analysis.
Other types of practices can be added in the future as further subclasses of \lstinline|DataPractice|, to provide a wide coverage of semantics.

To facilitate auditing, we retain the original text segment for each \lstinline|DataPractice|, to examine whether the representation is faithful or not. All \lstinline|DataPractice|s of a privacy policy belong to the same \lstinline|PrivacyPolicy| node, which also further links to the original website. As other types of online services/entities will also have privacy policies, we use a generic name \lstinline|Service|, to allow future extensions.

\section{Datasets}

\subsection{Privacy policy annotation dataset}
\label{sec:policy-ie-enrichment}

\begin{figure*}
   \centering
   \includegraphics[width=.9\linewidth]{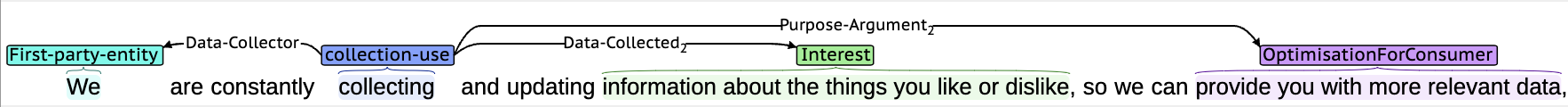}
   \caption{Sample annotation produced by annotators}
   \label{fig:annotation-sample}
\end{figure*}

Since no existing dataset satisfies our goal (Sec \ref{sec:background:pp-corpus}), to support the benchmarking, we enriched the Policy-IE dataset, by recruiting 2 domain experts (graduates with law-related degrees) to perform the annotation. Figure \ref{fig:annotation-sample} presents a sample annotation from our dataset.

The annotation took three phases, where we used the first phase (2 policies) to refine the guideline instructions for corner cases and allow the annotators to familiarize themselves with the procedure; the second phase (3 policies) verified their familiarity and consensus; they were then given 5 policies for the 3rd phase. The policies were selected alphabetically from Policy-IE. Starting from phase 1, each annotator received a detailed guideline, containing instructions on using the annotation tool, brat \cite{stenetorp_brat_2012}, instructions on tasks to complete, and descriptions for special cases (usually for handling uncertainties). The final dataset is formed after reconciliation.
The corresponding resource contains the full guidelines (\lstinline|Annotator-Instruction.pdf|), and the detailed annotation schema as brat configuration files.


In total, the annotators spent 1269min for annotating 10 privacy policies, leading to an enrichment of 2049 annotated entities, 129 relations, and 237 events. Among them, there are 694 fine-grained data and 438 fine-grained purpose annotations. As seen later, this shows sufficiency in testing model capabilities (as opposed to training a new model).

\begin{figure}
    \centering
    \includegraphics[width=.9\linewidth]{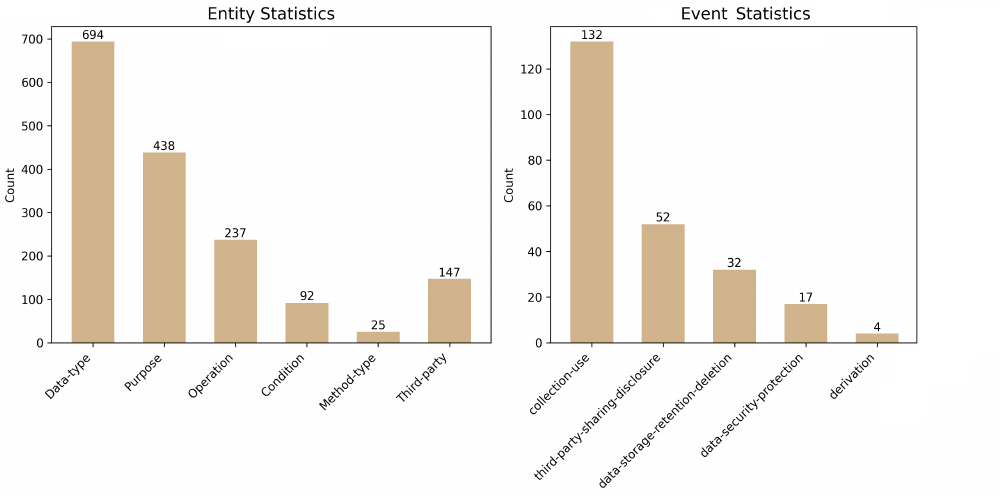}
    \caption{Statistics of entities and events in the annotation}
    \label{fig:annotation-entity-event}
\end{figure}

Diving deeper, Figure \ref{fig:annotation-entity-event} shows the distribution of different types of entities and events. We can observe that data and purposes are the most common entity types; there is a very limited number of ``condition'' (of data collection or storage) and ``method-type'' (data protection method). Likewise, ``collection-use'' and ``third-party-sharing-disclosure'' are the most common types of events, while the rest are less common. This builds our choice of focusing on  ``collection-use'' and ``third-party-sharing-disclosure'' in the NLP pipeline, as they are predominant and have an unambiguous evaluation strategy.

\subsection{\prprgraph{} for top-100 websites}

We used the NLP pipeline with the best-performing models to process the privacy policies of top-100 most visited websites, accordingly to 
Tranco list \cite{lepochat_tranco_2019}\footnote{We use the Tranco list \texttt{93VV2}, resembling most-visited websites between 3 July - 1 August 2024.}, and using privacy policies from the Princeton-Leuven Longitudinal Privacy Policy Dataset \cite{amos_privacy_2021}.\footnote{Because some websites do not have valid privacy policies in the dataset, we retry the next one until 100 policies are retrieved.} The results are saved as \prprgraph{}.

\begin{figure}
    \centering
    \includegraphics[width=\linewidth]{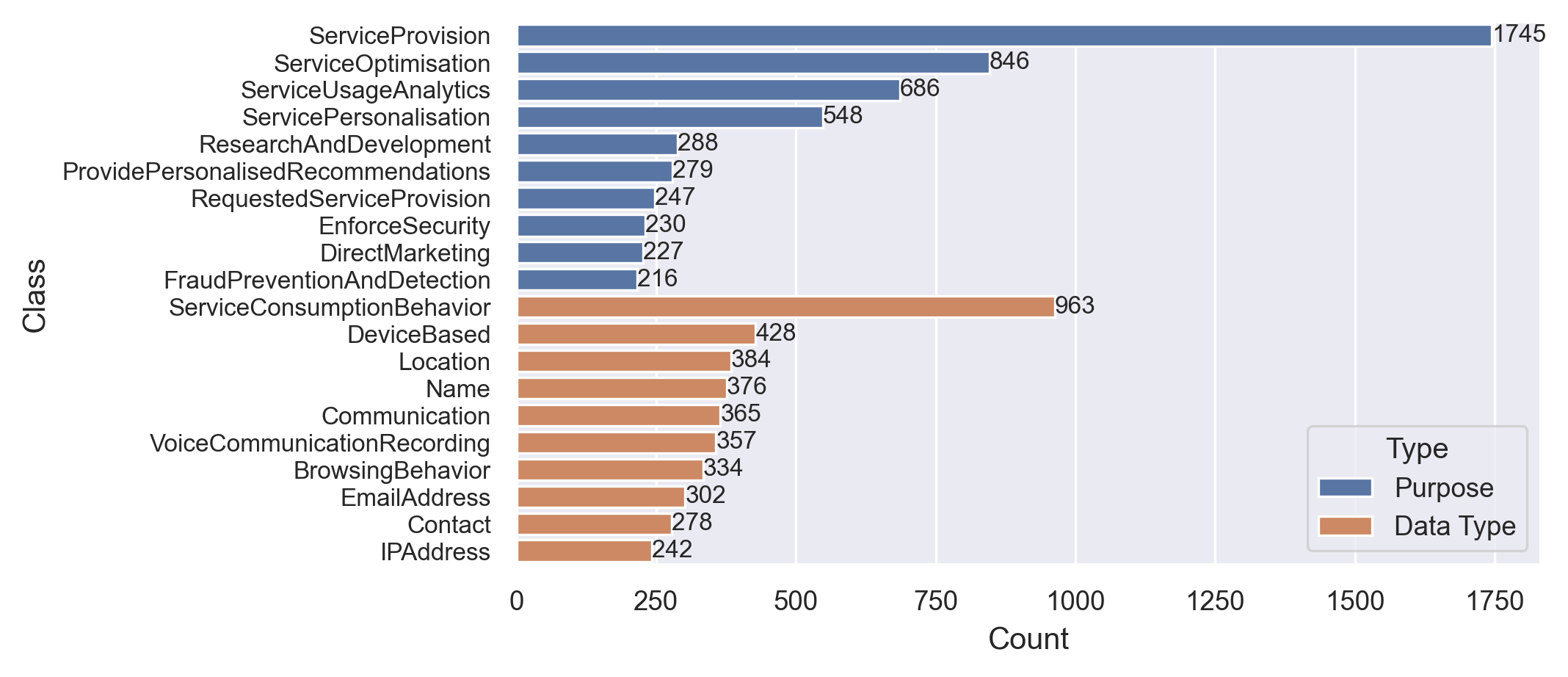}
    \caption{Top-10 classes for data and purpose, in \prprgraph{} of top-100 websites}
    \label{fig:kg-class-count}
\end{figure}

The resulting \prprgraph{} contains 84329 triples, of 11800 data privacy practices, where 6488 practices are explicit first-party data-collection-use, and 1324 practices are for third-party-sharing-disclosure. There are 10054 mentions of 128 distinct data classes, and 8743 mentions of 78 distinct purpose classes.
Figure \ref{fig:kg-class-count} shows the cumulative most common purposes and data types in \prprgraph{}.

\begin{figure}[htb]
   \centering
   \includegraphics[width=.9\linewidth]{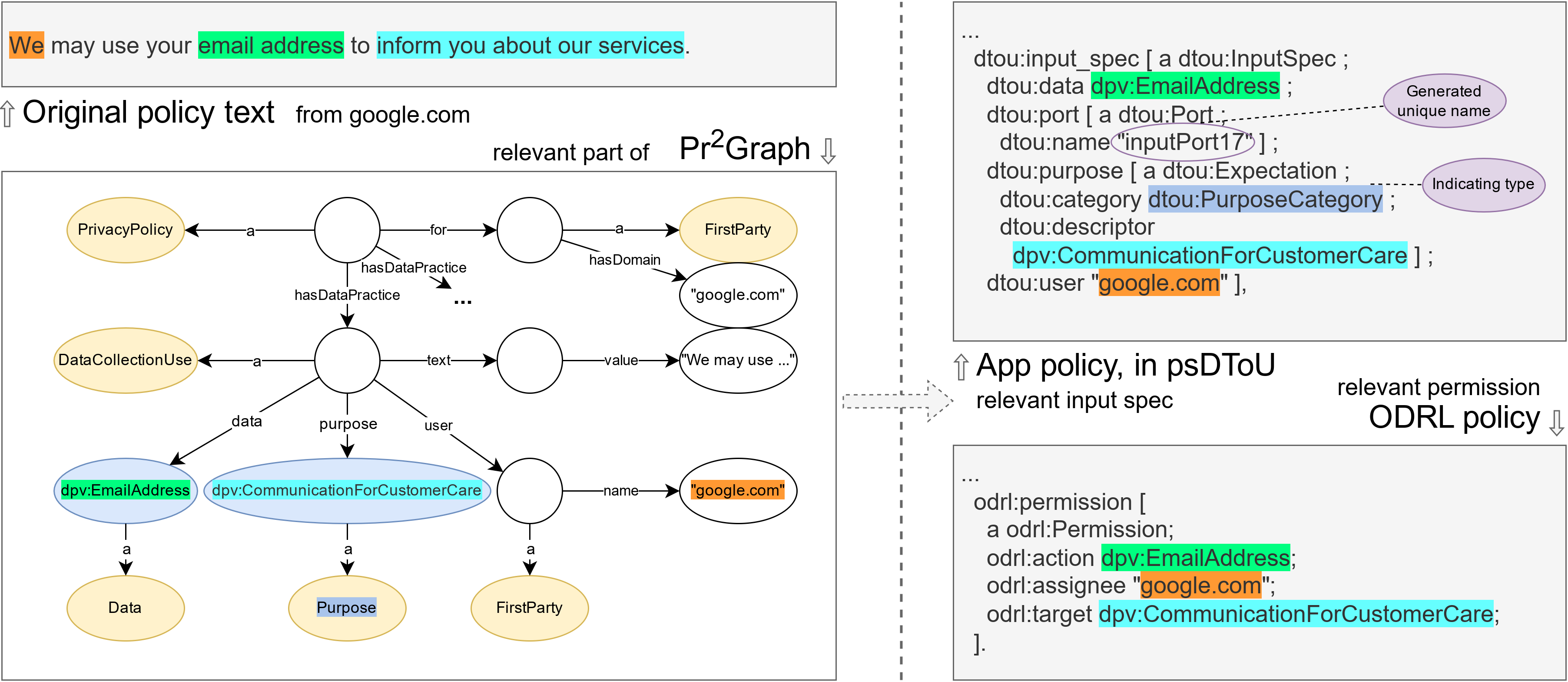}
   \caption{Conversion to \prprgraph{}, and further conversion to formal policy representation in ODRL and psDToU, using data class as data identifier for demonstration}
   \label{fig:dtou-convert}
\end{figure}

Different downstream tasks can be performed over \prprgraph{}. For example, one may construct actionable formal policies using a policy model, such as ODRL \cite{_odrl_2018} or \textit{app policy} in the psDToU language \cite{zhao_perennial_2024}.
Figure \ref{fig:dtou-convert} shows an example conversion, from natural-language policy text, to \prprgraph{}, and further to formal policies as ODRL and \textit{app policy}. Texts and entities of the same type are coloured consistently. We use the data type as data's identifier for demonstration, which may not fit every use case. The converted formal policies (both in ODRL and psDToU) are included in the resource for demonstration and facilitating reuse.


\section{Benchmark}

With the enriched annotation, we performed a series of experiments to evaluate LLMs' performance -- their accuracy (f1-score) in producing the correct outputs. In particular, we chose two main models, gpt-4o and gpt-4o-mini, together with their fine-tuned versions; we also included other models for comparison, especially o1 and o3-mini\footnote{More specifically, we used gpt-4o-mini-2024-07-18 (\lstinline|mini|), gpt-4o-2024-08-06 (\lstinline|4o|), gpt-4o-2024-11-20 (\lstinline|4o-Nov|) (unavailable for fine-tuning), o1-2024-12-17 (\lstinline|o1|), and o3-mini-2025-01-31 (\lstinline|o3-mini|), which were the latest state-of-the-art models at the point of the benchmarking.}. The experiments executed queries under different configurations, and calculated the f1-scores.

\subsection{Experimental setting}

\paragraph{Tasks}
We fixed the hyperparameters, especially \lstinline|temperature = 0|, to obtain stable outputs.
We evaluated the original models, as well as shallow-fine-tuned models using a small portion of data evenly sampled from our dataset\footnote{We also performed few-shot prompting, but discovered lower performance, and thus switched to fine-tuning.}.
For each model, each pipeline step, we execute the query and obtain the query results. We iteratively refined the query details and fine-tuning data size selection. 

\paragraph{Shallow fine-tuning}
For the fine-tuning, we only used a small portion of the dataset as training data (thus calling it shallow fine-tuning) -- among all strategies, the maximum is 120 (out of 1087 data points). We intentionally limited the portion of data, contrary to usual practices when training new models, due to our different goal: to evaluate the capability of existing LLMs. There are also two practical reasons for performing fine-tuning: 1) the LLM output should comply with our specified schema (which is often otherwise before fine-tuning); and 2) the LLM should understand some nuanced preferences hard to describe through instructions.

\paragraph{Fine-tuning data selection}
When selecting fine-tuning data, we distinguish between and use both ``non-empty'' data and ``empty'' data -- ``non-empty'' data means data points with non-empty targeting annotations, while ``empty'' data means otherwise. 
We iteratively and heuristically determined a few sets of data sizes\footnote{For example, ``10-30-2-6'' denotes 10 non-empty training data, 30 empty training data, 2 non-empty validation data, and 6 empty validation.}, namely 10-30-2-6, 20-20-4-4, and 40-80-10-20.


\subsection{Metrics}

For most tasks, we calculate the F1 score as a common practice for balancing between precision and recall, following this formula:
\begin{align*}
   &precision = tp / (tp + fn) \\
   &recall = tp / (tp + fp) \\
   &F1 = 2 \cdot precision \cdot recall / (precision + recall)
\end{align*}
where $tp$ indicates true positive, $fn$ indicates false negative, $fp$ indicates false positive.

For tasks with word matching, we use a \emph{relaxed} metric where the predicted result (text span) is proportionately considered in scoring -- if the longest-common-substring ($lcs$) ratio of the (predicted and expected) result is over a given threshold ($0.9$), the true positive is increased by $lcs$; otherwise, only exact matching is considered true positive. This is to facilitate cases where the model output and correct output only misses a few non-functional words (e.g.~``the'') and being considered false match.

In addition to the overall macro f1-score (f1) for each task, we include two detailed facets of the scores: f1-non-empty (f1-n) denotes the macro f1 over data that should not be predicted empty, and f1-empty (f1-e) denotes the macro f1 over data that should be predicted empty. These two additional scores help us better understand the LLM performance in different input situations, and is a factor to consider in model selection.

\subsection{Results}

\begin{table*}[h]
    \centering
    \includegraphics[width=\linewidth]{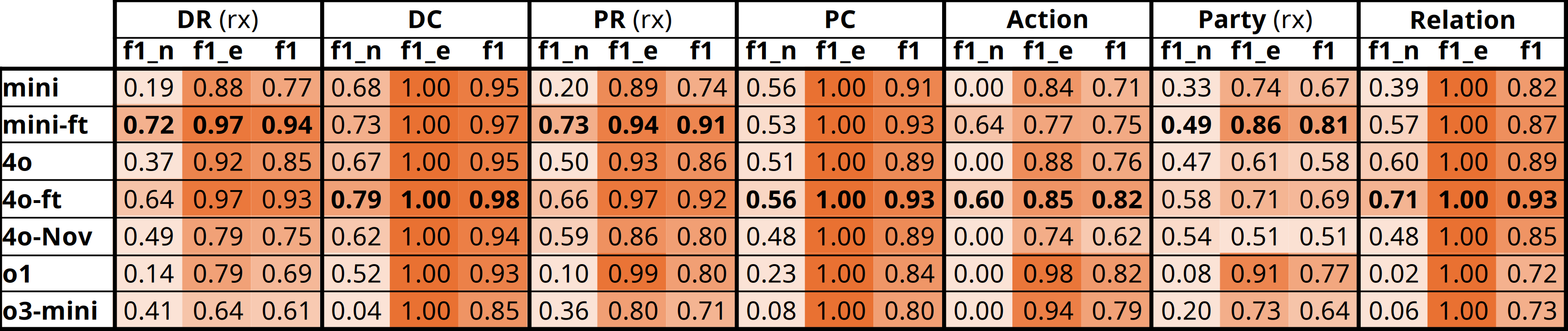}
    \caption{LLM model performance in evaluation. DR means Data Recognition; DC means Data Classification; PR means Purpose Recognition; PC means Purpose Classification; Action, Party and Relation means the corresponding recognition job. \lstinline|f1_n| refers to f1-non-empty metric; \lstinline|f1_e| refers to f1-empty metric; \lstinline|f1| refers to macro f1 metric. \lstinline|rx| indicates \emph{relaxed} matching. Models end with \lstinline|-ft| are fine-tuned models.}
    \label{tab:llm-eval-result}
\end{table*}

Table \ref{tab:llm-eval-result} summarizes the main results of the best-performing settings from our experiments.
Each line denotes a model, and each column group denotes a different task in our pipeline. As can be seen, base models have different performance across different tasks, while fine-tuning generally improves the performance (though not always). In addition, bold texts indicate the model we chose when analyzing the top-100 websites.

Overall, for the best-performing models, most tasks have f1-score of about 0.9 or higher. This means the models are able to correctly identify the desired entities in most tasks. In particular, the very high f1-empty scores indicate that the models are particularly good at determining whether the given segment contains targeting information or not. This means that the model will unlikely produce non-existent entities for the privacy policy, reducing the worry about hallucination.

On the other hand, when focusing on the scores for only non-empty-valued segments (f1-n), the score becomes lower, to around 0.5 - 0.7. This indicates that the models are not always accurate in labeling the entities, and still have space to improve. Nevertheless, the performance of the models does not indicate they are unacceptably worse than human annotators, as the inter-annotator agreements are also not perfect (e.g.~that for data classification grew from 55\% to 72\% to 85\% in the three phases for our dataset), and the model performance is, despite lower, reasonably comparable to that.

We also notice that bigger models (e.g.~gpt-4o) is not necessarily better than smaller models (e.g.~gpt-4o-mini), nor fine-tuning is always better, such as from that in purpose recognition and purpose classification. Also, sometimes the model was already good at the tasks (esp.~data and purpose classification) without fine-tuning. This potentially indicates that detailed and targeted prompts in job descriptions may already be enough for instructing the model. In the meantime, the reasoning models appeared less appropriate for our tasks, given their generally lower accuracy for most tasks.

Overall, our result shows evidence that state-of-the-art LLMs are able to produce meaningful predictions for identifying information for data practices with shallow fine-tuning, reaching performance comparable to human annotators. This shows confidence in using them for annotating privacy policies instead of human annotators, fostering faster and more cost-efficient annotations, thus featuring auditability and reliability for downstream tasks.

\section{Resource maintenance, limitation and future work}

We continuously refine the procedure, tools, and the dataset. For example, the prompting strategy for data entity extraction reported in this paper is the 4th version.
So far, we have made the source code of the whole pipeline and all curated and produced datasets available, to allow anyone to run the same or modified experiments and produce more datasets, if they wish to contribute.

We aim to test and develop the general NLP pipeline further, such as by further benchmarking other LLMs or improving performance. A dashboard would be useful to better illustrate the results.
With that, we will analyze more privacy policies and update the datasets.

Despite being 10 times cheaper than human annotators, about \$2.2 is still required for analyzing each privacy policy using our pipeline (\$888.32 in total during our development and deployment)\footnote{For the human annotators, about \$10 - \$20 is required for annotating one privacy policy.}. This constitutes the reason for releasing public datasets of the analysis results.
We are also looking for appropriate methods to reduce the cost burden, such as using local models, as this is a perpetual task until services provide formal policy descriptions.

Since the current pipeline focuses on data collection and data sharing, due to the limited availability of ground-truth data and the complexity of details of other event types, future work should collect more expert annotations, and explore mechanisms to evaluate LLM's capability for other event types for better coverage of semantic information.

\section{Summary}

In this paper, we presented a semantic-centric approach and corresponding resources for filling in a critical gap in the source of formal policies for tasks like automated compliance analysis. We explained the design of the NLP pipeline, the structure of \prprgraph{}, our constructed annotation dataset for evaluating the NLP pipeline, and further obtained \prprgraph{} for the top-100 most-visited websites, as well as demonstrating downstream usages for constructing formal policies; we also showed the benchmark design and results, verifying the capability of the NLP pipeline. All resources are made public for reuse and reference.

We believe the pipeline presented in this paper enables the opportunity of easily accessible tools for analyzing privacy policies. The constructed \prprgraph{} also serves as a source for downstream task usage, such as providing much-needed format policy specifications for research, and automated agents verifying compliance of data usage practices and/or making decisions. Together, they form a solid basis for semantic-powered interoperable approaches for auditing the Web and supporting users' decision making.



\bibliographystyle{splncs04}
\bibliography{cited_only}

\appendix
\section{LLM prompting strategy}
\label{appendix:llm-prompting-strategy}

\begin{figure}
   \centering
   \includegraphics[width=\linewidth]{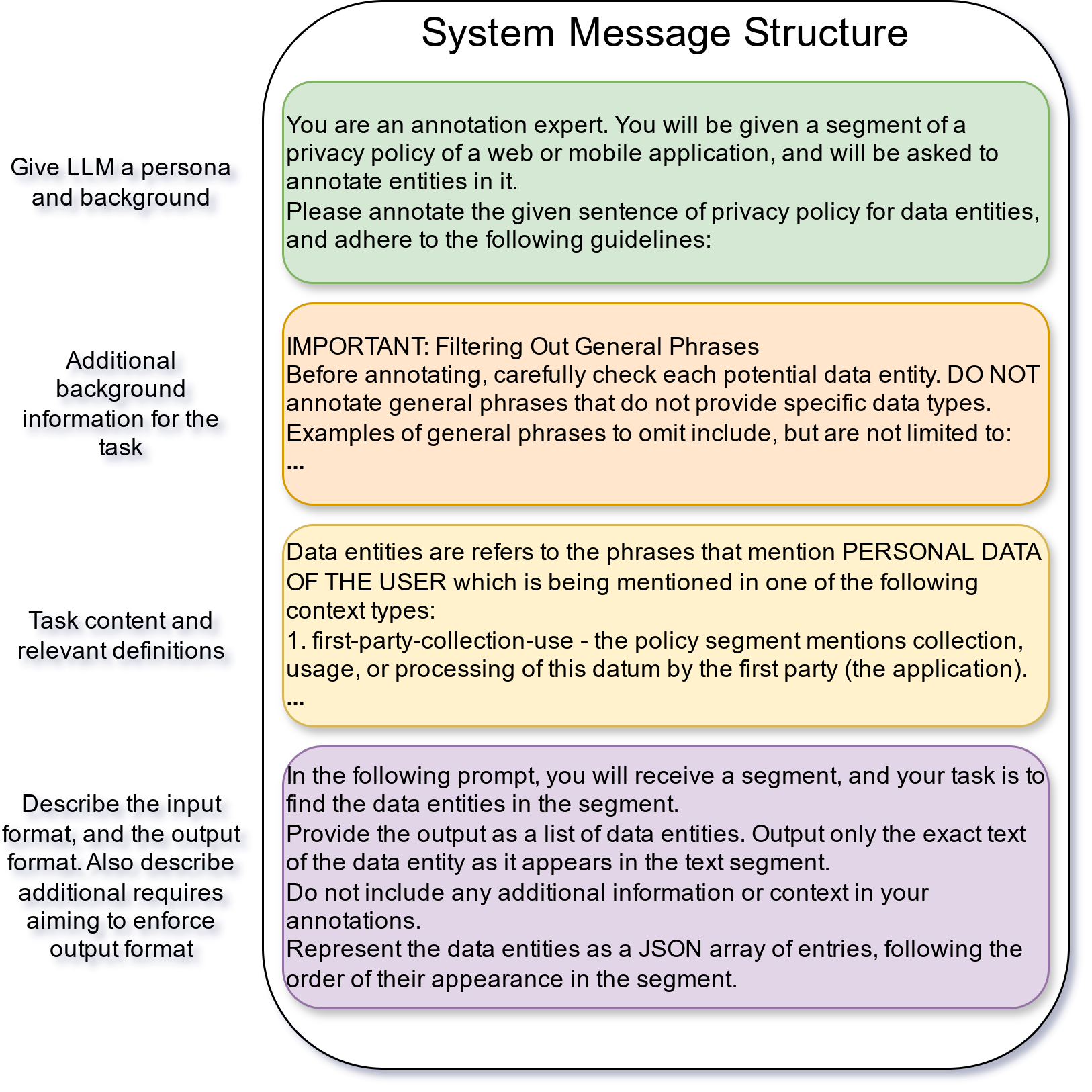}
   \caption{General structure of the system message of our LLM prompts}
   \label{fig:llm-prompt}
\end{figure}

Figure \ref{fig:llm-prompt} shows the general structure of the system messages (where each task has task-specific modifications), which follows a prompting strategy similar to those found in related literature. For some tasks (e.g.~data entity classification), there are multiple parts in the input, such as the \emph{segment} and \emph{data entities} to classify. In such cases, the user message has explicit marks of the boundary, and this is also explained in the system message.

\end{document}